\pdfoutput=1

\documentclass[11pt]{article}

\usepackage[]{EMNLP2023}

\usepackage{times}
\usepackage{latexsym}

\usepackage[T1]{fontenc}

\usepackage[utf8]{inputenc}

\usepackage{microtype}

\usepackage{inconsolata}

\usepackage{amsfonts}
\usepackage{amsmath}
\usepackage{mathtools}
\usepackage{xfrac}
\usepackage{amsthm}
\usepackage{xspace}

\usepackage{booktabs}

\usepackage{float}

\usepackage{graphics}
\usepackage{makecell}

\usepackage{xcolor}

\usepackage{graphicx}
\usepackage{subcaption}
\usepackage{adjustbox} 

\usepackage{hyperref}
\usepackage[nameinlink]{cleveref}

\newcommand{\Audio}{\ensuremath{A}\xspace}
\newcommand{\Text}{\ensuremath{T}\xspace}
\newcommand{\Multimodal}{\ensuremath{M}\xspace}
\newcommand{\hidden}{\ensuremath{h}\xspace}
\newcommand{\clf}{\ensuremath{CLS}\xspace}

\newcommand{\whisbert}{WhisBERT\xspace}
\newcommand{\babylm}{BabyLM\xspace}

\title{
\whisbert: Multimodal Text-Audio Language Modeling on 100M Words
}

\usepackage[safe]{tipa}

\newcommand{\ethz}{\text{\normalfont \textipa{Q}}}

\newcommand{\mittech}{\normalfont \text{\textipa{M}}}  
\newcommand{\stanford}{\normalfont \text{\textipa{S}}}

\author{
Lukas Wolf$^{\ethz}$\quad
Greta Tuckute$^{\mittech}$ \quad 
Klemen Kotar$^{\stanford}$ \quad 
Eghbal Hosseini$^{\mittech}$ \\ 
\textbf{Tamar I. Regev$^{\mittech}$}   \quad 
\textbf{Ethan Gotlieb Wilcox$^{\ethz}$}\quad 
\textbf{Alex Warstadt$^{\ethz}$} \\
$^{\ethz}$ETH Z\"{u}rich \quad
$^{\mittech}$MIT \quad
$^{\stanford}$Stanford University \\
{
\tt{\{\href{mailto:wolflu@ethz.ch}{wolflu}, \href{mailto:warstadt@ethz.ch}{warstadt}, \href{mailto:ethan.wilcox@ethz.ch}{ethan.wilcox}\}}@ethz.ch} \quad \\\tt{\href{mailto:klemenk@stanford.edu}{klemenk@stanford.edu}} \quad
\tt{\{\href{mailto:gretatu@mit.edu}{gretatu},
\href{mailto:ehoseini@mit.edu}{ehoseini}, \href{mailto:tamarr@mit.edu}{tamarr}\}@mit.edu} \quad
}

\begin{document}
\maketitle
\begin{abstract}
Training on multiple modalities of input can augment the capabilities of a language model. Here, we ask whether such a training regime can improve the \emph{quality} and \emph{efficiency} of these systems as well. We focus on text--audio and introduce \whisbert, which is inspired by the text--image approach of FLAVA \citep{singh_flava_2022}. In accordance with \babylm \citep{warstadt2023papers} guidelines, we pretrain \whisbert on a dataset comprising only 100 million words plus their corresponding speech from the word-aligned version of the People's Speech dataset \citep{galvez_peoples_2021}. 
To assess the impact of multimodality, we compare versions of the model that are trained on text only and on both audio and text simultaneously. We find that while \whisbert is able to perform well on multimodal masked modeling and surpasses the \babylm baselines in most benchmark tasks, it struggles to optimize its complex objective and outperform its text-only \whisbert baseline.

\vspace{0.5em}
\includegraphics[width=1.25em,height=1.25em]{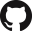}\hspace{1em}\parbox{\dimexpr\linewidth-2\fboxsep-2\fboxrule}{\url{https://github.com/lu-wo/whisbert}}

\end{abstract}

\section{Introduction}

Recent advances in language modeling and their downstream applications have been driven, in large part, by bigger models, with respect to both model size and amounts of training data. Larger and larger pretraining datasets highlight the gap between humans and deep learning models in terms of learning efficiency ---while state-of-the-art language models need billions of examples to approach human-level language performance, people learn their language from experience with about 100 million words or less \cite{warstadt2022what,frank2023bridging}.

We hypothesize that one major reason for this data efficiency gap is the different inputs that humans and current deep learning systems receive. Human language learning involves multiple modalities, including both visual and auditory input. In contrast, typical language models are trained on representations of text alone. For this \babylm submission, we ask whether training on inputs of multiple modalities can increase language models' training efficiency, with a focus on text-audio multimodal input. We conjecture that multimodal data sources have the potential to enrich the language learning process, enabling models to leverage complementary information from different modalities and thus augment their learning capacity \cite{baltrušaitis2017multimodal}.

Multimodal language modeling has experienced a noteworthy surge in research productivity lately, in applications such as image retrieval, semantic embeddings,  and image generation \cite{driess2023palm, koh2023generating, yasunaga2023retrieval}
However, text-audio multimodal language modeling (e.g. \cite{chuang2019speechbert, lakhotia2021generative}) remains largely unexplored, especially in low-resource settings such as the 100 million training regime we employ here. As a first step towards a text-audio language model, we introduce \whisbert, a novel masked language model (MLM) architecture inspired by vision-text models such as FLAVA \citep{singh_flava_2022}. The core idea is that \whisbert is trained in a multitask setting on both unimodal (i.e. text- or audio-only) and multimodal objectives. In multimodal objectives, the model receives matched text-audio segments, and it can use information from one modality to learn representations for the other.

To accommodate the specific requirements of the \babylm challenge \cite{warstadt2023papers}, we pretrain WhisBert on a dataset of matched audio and text transcripts comprising 100 million words sampled from the People’s Speech dataset \cite{galvez_peoples_2021}. 
We use an improved version of the audio-text-aligned training data, a subset of an upcoming speech production dataset release (see \Cref{sec:dataset}). This commitment to using high-quality pretraining data is in line with the data efficiency objectives of the \babylm challenge.

We carry out a rigorous evaluation of the performance of the audio, text, and multimodal encoders within this new framework. 
We find that even though the optimization problem in the multimodal setting is much harder compared to a unimodal setting, the multimodal \whisbert model outperforms the text-only baseline in a majority of the \babylm challenge tasks, which address several aspects of language understanding, even when trained for only a single iteration over the dataset.

\section{\whisbert}
\label{sec:methods}

\whisbert is a multimodal audio and text model that is inspired by \textit{OpenAI}'s Whisper model \cite{radford2022robust} for speech recognition and BERT \cite{devlin2019bert} for bidirectional language encoding. \whisbert contains two separate input streams, one of audio and of its corresponding text (i.e., a transcription). The model is trained using a combination of two unimodal and three multimodal masked training objectives. In the unimodal setting, the model must predict either a masked word or a masked patch of audio. In the multimodal training setting, the model must predict pairs of matched word/audio patches. This multi-objective training setup is inspired by the visual-audio model FLAVA \cite{singh_flava_2022}.

\subsection{Architecture details}
\label{sec:architecture_details}

\paragraph{Audio encoder}
To create audio patches that we can process with Whisper's bidirectional transformer encoder \cite{vaswani2017attention}, the audio stream is first passed through the Whisper Feature Extractor available on \href{https://huggingface.co/docs/transformers/model_doc/whisper}{Hugging Face}.

All audio data is re-sampled to a rate of 16,000 Hz, and an 80-channel log-magnitude Mel spectrogram representation is computed using 25-millisecond windows with a 10-millisecond stride. 
We then pass the audio spectrogram through a patch embedding layer: a convolutional encoder processes the extracted frequency features using a stem of two 1-dimensional convolution layers (along the time dimension, filters cover all input frequencies), both with a filter width of 16 and incorporating the GELU activation function. The second convolution layer employs a stride of 10. This patch embedding layer creates overlapping 1-dimensional audio patches covering 100ms of the audio signal as input to the transformer.

After preprocessing and patch embedding, sinusoidal position embeddings are added to the stem's output, which is then processed by Whisper's transformer encoder blocks. A notable difference to the standard Whisper encoder is that we prepend a learnable classification (henceforth, \clf) token at the beginning of the audio patch sequence. Therefore, the audio encoder produces a list of audio hidden states $\{\hidden_\Audio\}$ each corresponding to a contextualized audio patch, as well as an additional audio classification state $\hidden_{\clf, \Audio}$.

\paragraph{Text encoder}
In order to encode the text input, we choose a standard bidirectional transformer architecture following the BERT \cite{devlin2019bert} model. We train a WordPiece \cite{wu2016googles} tokenizer on the 100M words in our People's speech \cite{galvez_peoples_2021} subset (see \Cref{sec:dataset}). The WordPiece tokenizer automatically prepends a \clf token to the token sequence which is contextualized with the rest of the sequence. The text encoder produces a list of text hidden states $\{\hidden_\Text\}$ corresponding to a text token, as well as an additional text \clf token $\hidden_{\clf, \Text}$.

\paragraph{Multimodal encoder}
The multimodal encoder is a standard transformer encoder that gets as input the concatenated contextualized audio and text sequences. Additionally, we prepend a learnable multimodal \clf token and employ sinusoidal positional embeddings. The multimodal encoder contextualizes the multimodal sequence and  outputs a list of multimodal hidden states $\{\hidden_\Multimodal\}$ each corresponding to an unimodal vector from $\{\hidden_\Audio\}$ or $\{\hidden_\Text\}$, as well as an additional multimodal \clf token $\hidden_{\clf, \Multimodal}$.

\paragraph{Adapting to downstream tasks}
The WhisBert model can be readily applied to both unimodal and multimodal tasks. For audio recognition tasks (e.g., speaker identification or speech recognition), we apply a classifier head (e.g., a linear layer or a multi-layer perceptron) on top of the unimodal classification token, \(\hidden_{\clf, \Audio}\), from the audio encoder. Similarly, for language understanding and multimodal reasoning tasks, we can apply a classifier head on top of the classification token, \(\hidden_{\clf, \Text}\), from the text encoder or \(\hidden_{\clf, \Multimodal}\) from the multimodal encoder, respectively.

\subsection{Pretraining objectives}
\label{sec:training_objectives}
Our goal is to pretrain models to have robust contextual representations for both text and audio on their own as well as for aligned text-audio pairs. We use the approach from FLAVA \cite{singh_flava_2022} of multitask training over a selection of unimodal and multimodal training objectives that have been demonstrated to facilitate joint learning on images and text. We adapt the five objectives used by FLAVA for the audio domain. 

\subsubsection{Unimodal pretraining objectives}
\label{sec:unimodal_objectives}

\paragraph{Masked Language Modeling}
\label{sec:masked_language_modeling}
Masked Language Modeling (MLM) is a pretraining objective that encourages the model to learn a deep understanding of the language. In MLM, a portion of the input tokens is masked and the model is trained to predict the original identity of the masked tokens based on their context. 

Given an input sequence of tokens \(x = [x_1, x_2, ..., x_T]\), for MLM, a subset \(M\) of these tokens is selected to be masked. The objective is to minimize the negative log-likelihood of the masked tokens:
\begin{equation}
L_{\text{MLM}}(x) = -\frac{1}{|M|} \sum_{t \in M} \log p_{\text{model}}(x_t | x_{\neg t})
\end{equation}

Here, \(x_t\) is a masked token, \(x_{\neg t}\) represents the sequence with the token \(x_t\) masked, and \(p_{\text{model}}\) is the model's probability distribution over possible tokens. \(|M|\) is the size of the subset of masked tokens, and the sum is taken over all masked positions \(t\). The goal is to adjust the model's parameters to minimize this loss. We obtain a probability distribution over the vocabulary by applying a linear prediction head on the text hidden states $\{\hidden_\Text\}$.

\paragraph{Masked Audio Modeling}
\label{sec:masked_audio_modeling}
We introduce the Masked Audio Modeling (MAM) objective $L_{MAM}$ which follows the principles of Contrastive Predictive Coding \cite{oord2019representation}. In MAM, we randomly mask audio patches in the input sequence to the audio encoder. The encoder is expected to generate outputs that are most similar to the unmasked input at a particular masked position $t$. The self-supervised loss function, which aims to encourage the model to align masked tokens with their unmasked identities given the context, is defined for a masked token localized at $t$ as:
\begin{equation}
L_{\text{MAM}} = -\log \frac{\exp(sim(c_t, b_t)/\kappa)}{\sum_{b_i \in B_D} \exp(sim(c_t, b_i)/\kappa)}
\end{equation}

Here, \(c_t\) is the output of the transformer at position \(t\), and \(b_i\) is the audio representation vector of the (unmasked) patch at some offset \(i\). \(B_D\) is a set of 20 uniformly selected negative samples from the same sequence, plus \(b_t\), and $sim()$ is a similarity function. For our implementation, we use the cosine similarity function, adjusted by a temperature function, $\kappa$, which is set to $0.1$.
The loss function operates by adjusting the output of the transformer at position \(t\) to be most similar to the encoded representation at \(t\), despite the fact that this input to the transformer is masked. In this way, the model is encouraged to predict the content of the masked spans based on the unmasked context.

\subsubsection{Multimodal Pretraining Objectives}
\label{sec:multimodal_objectives}

\paragraph{Multimodal Contrastive Loss}
Contrastive loss \cite{gutmann2010contrastive} has been successfully applied to image-text representation learning in approaches such as CLIP \cite{radford2021clip}. 
Our audio-text contrastive loss $L_{MMC}$ aims to maximize the cosine similarities between matched audio and text pairs and minimize those for the unmatched pairs across a given batch of audio clips and corresponding text. This is achieved by linearly projecting the classification token of each audio sequence $\hidden_{\clf, \Audio}$ and text sequence $\hidden_{\clf, \Text}$ into a common embedding space, followed by L2-normalization, dot-product, and a softmax loss scaled by temperature.

The goal of this process is to ensure that the audio and text representations for the same data point are brought closer together in the embedding space, while representations for different data points are pushed apart. This encourages the model to learn meaningful representations that capture the shared information between the audio and text modalities.

\paragraph{Masked Multimodal Modeling (MMM)}
We introduce a Masked Multimodal Modeling (MMM) pretraining objective $L_{MMM}$, that uses the output of the multimodal encoder $\{\hidden_\Multimodal\}$ to attempt to reconstruct the masked tokens from both the audio and text sequences. For the multimodal contextualized audio tokens, we employ the Contrastive Predictive Coding strategy introduced in \Cref{sec:masked_audio_modeling}. For the multimodal text tokens, we add a multimodal masked language modeling head we compute the MLM loss as introduced in \Cref{sec:masked_language_modeling}. 

The MMM pretraining objective is designed to encourage the model to understand the interdependencies between audio and text modalities, which in addition to the MMC loss has been found to improve performance on multimodal downstream tasks \cite{singh_flava_2022}. It is computed separately from the contrastive loss, which is applied on audio and text tokens without any masking.

\paragraph{Audio-Text Matching (ATM)}
Finally, we incorporate an Audio-Text Matching loss, $L_{ATM}$, in which we feed a batch of samples that include both matched and unmatched audio-text pairs.  We apply a classifier on top of the output from the multimodal encode to decide if the input audio and text match each other.

\begin{table*}[ht]
\centering
\scalebox{0.8}{
    \begin{tabular}{|l|c|c|c|c|c|}
    \hline
    Task & MLM & MM & OPT-125m & RoBERTa-base & T5-base \\
    \hline
    anaphor\_agreement & 83.74\% & 81.29\% & 63.8\% & 81.5\% & 68.9\% \\
    argument\_structure & 68.60\% & 64.88\% & 70.6\% & 67.1\% & 63.8\% \\
    binding & 66.95\% & 65.38\% & 67.1\% & 67.3\% & 60.4\% \\
    control\_raising & 65.25\% & 64.76\% & 66.5\% & 67.9\% & 60.9\% \\
    determiner\_noun\_agreement & 92.24\% & 87.93\% & 78.5\% & 90.8\% & 72.2\% \\
    ellipsis & 83.14\% & 88.68\% & 62.0\% & 76.4\% & 34.4\% \\
    filler\_gap & 73.12\% & 72.02\% & 63.8\% & 63.5\% & 48.2\% \\
    irregular\_forms & 89.62\% & 85.90\% & 67.5\% & 87.4\% & 77.6\% \\
    island\_effects & 53.51\% & 55.87\% & 48.6\% & 39.9\% & 45.6\% \\
    npi\_licensing & 64.77\% & 55.12\% & 46.7\% & 55.9\% & 47.8\% \\
    quantifiers & 69.58\% & 71.69\% & 59.6\% & 70.5\% & 61.2\% \\
    subject\_verb\_agreement & 75.05\% & 70.73\% & 56.9\% & 65.4\% & 65.0\% \\
    hypernym & 50.12\% & 51.98\% & 50.0\% & 49.4\% & 48.0\% \\
    qa\_congruence\_easy & 71.88\% & 67.19\% & 54.7\% & 31.3\% & 40.6\% \\
    qa\_congruence\_tricky & 52.12\% & 53.94\% & 31.5\% & 32.1\% & 21.2\% \\
    subject\_aux\_inversion & 77.90\% & 74.85\% & 80.3\% & 71.7\% & 64.9\% \\
    turn\_taking & 61.79\% & 58.21\% & 57.1\% & 53.2\% & 45.0\% \\
    \hline
    \end{tabular}
}
\caption{Evaluation scores of text-only (MLM), multimodal WhisBERT (MM), and the \babylm baselines on \textit{BLiMP} tasks. The \babylm baselines were trained on the 100M words \babylm dataset.} 
\label{tab:extended_table}
\end{table*}

\subsection{Pretraining \whisbert}
\label{sec:pretraining-whisbert}
We pretrain \whisbert on both text and audio samples from the dataset introduced in \Cref{sec:dataset} for five epochs with stochastic gradient descent. Although \whisbert is able to learn both from paired and unpaired examples, in our pretraining dataset we only encounter text-audio pairs. This allows us to always apply all unimodal and multimodal objective functions. For further details and hyperparameters we refer to \href{https://github.com/lu-wo/whisbert}{this GitHub repository.}

\section{People's Speech Dataset}
\label{sec:dataset}
The People's Speech dataset \cite{galvez_peoples_2021} is a free-to-download, 30k hour English speech recognition dataset. The dataset is collected from appropriately licensed internet audio data with existing transcriptions, consisting of a clean and a dirty subset. We re-transcribed and re-aligned the People's Speech dataset using recently-released automatic speech recognition toolkits \citep{radford2022robust, bain2023whisperx}, which may provide better alignment than the baseline, publically available alignments.
For this step we transcribe speech the Whisper large-v2 model from OpenAI \cite{radford2022robust}. Numerals and non-standard characters were suppressed in the transcriptions, such that numbers were represented as words and non-standard characters were omitted. Otherwise, default parameters were used. The transcriptions were force-aligned to match the audio files using the WhisperX pipeline \cite{bain2023whisperx, Bredin2019PyannoteAudioNB, Baevski2020wav2vec2A}. We excluded very short transcripts (fewer than 100 words) or transcripts that contained more than 0.1\% of words that could not be transcribed. The remaining files were sorted according to mean word-level transcription confidence (Whisper estimates a value between 0 and 1 that denotes the transcription confidence per word). We selected the files containing the first 100M words in this ordering. The average confidence of these final 100M words was 0.78 with 47M words from the clean audio subset and 53M words from the dirty audio subset. The transcribed, word-aligned dataset will be released as part of an upcoming speech production dataset.

\section{Experimental Results}
\label{sec:results}

The main question we are interested in is whether pretraining on audio--text data can improve model performance. We assess this by comparing the text-encoder only version of \whisbert compared to the exact same architecture trained with the multimodal objectives introduced in \Cref{sec:training_objectives}. (This is the MLM (text) vs. MMM (multi-modal) comparison in \Cref{tab:extended_table}.) Our results suggest that the answer is mixed. The MLM (text-only) version of the model achieves higher scores on 12 out of the 17 test suites, with the multi-modal model performing higher for Ellipsis, Island Effects, Quantifiers, Hypernym, and Question/Answer Congruence (tricky) tests. Interestingly, the three of these that were in the original BLiMP paper (Ellipsis, Island Effects and Quantifiers), were three of the four lowest-scoring tests for human accuracy, suggesting that where multi-modality \emph{does} help, it is in processing particularly syntactically difficult material. Both of our trained models outperform the OPT-125M, RoBERTa and T5 baselines, averaging across tasks.

\begin{figure}
    \centering
    \scalebox{0.34}{
    \includegraphics{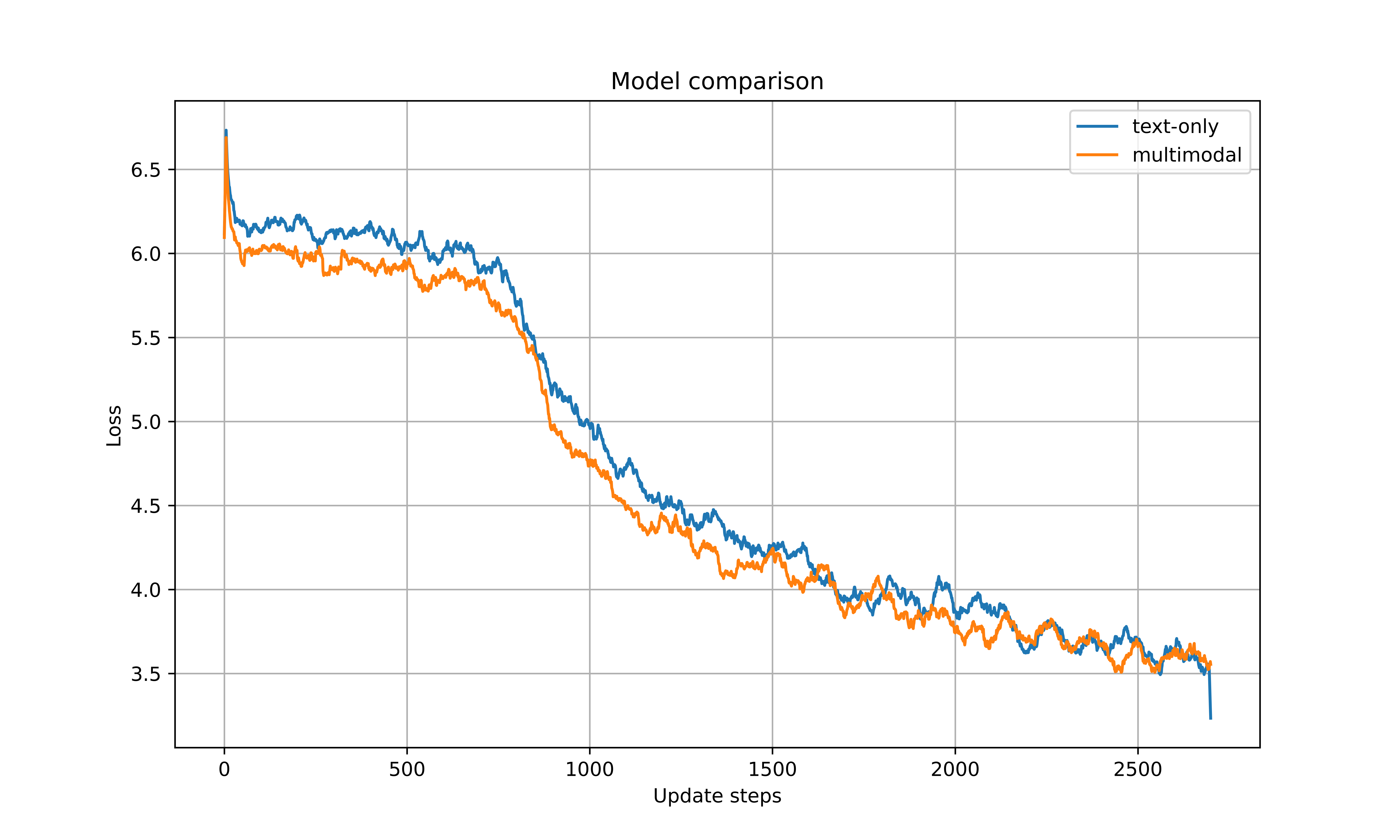}
    }
    \caption{Text-only baseline vs \whisbert on masked language modeling task during the first epoch. Interestingly, during the first epoch \whisbert seems to perform better (outperforming the text-only baseline in 11 out of 17 tasks), but after five epochs does not outperform the text-only baseline across all benchmark tasks}
    \label{fig:text_vs_multimodal}
\end{figure}

\section{Discussion}
\label{sec:discussion}

\paragraph{Limitations}

We begin our discussion by noting the limitations of the current work. First, the People's Voice dataset presents a unique set of challenges, which likely resulted in limitations of the \whisbert model. The most significant of these is that it is primarily comprised of audio from movies, and thus includes things like background noise, music and audio effects that accompanied the dialog. This could have resulted in lower text--audio alignment accuracy, and likely made the audio-modeling challenge more difficult than for an in-studio recorded dataset. 

Second, the requirements of the BabyLM challenge presented us with additional restrictions. Most notably, we were not allowed to use pretrained audio encoders, and thus had to train these from scratch. Likely, this contributed
to sub-optimal performance and requires further exploration. Furthermore, due to time limitations, we did not fully explore the space of the model's hyperparameters; it is well known that changes in hyperparameter settings can have large impacts on a model's performance. 

Our mixed results when comparing \whisbert against a text-only model suggest that small data settings are insufficient for effectively training a text-only masked language model. Given that the architectural basis for \whisbert, Flava, was designed and built as a large-data foundation model, we suggest that such larger-data settings serve as the basis for future development and testing of the \whisbert model.

\paragraph{Future Work}
We plan to train versions of \whisbert on more than 100M words and their corresponding audio. This would enable investigations of the full capacity of the \whisbert model and make it more comparable to similar vision-text models such as FLAVA \cite{singh_flava_2022}. On the architecture level, one could replace the bidirectional transformer in the \whisbert architecture with an autoregressive language model, allowing the use of the standard Whisper pretraining objectives in addition to the multi-modal ones.  

\section*{Contribution Statement}

LW, EH, TIR, EGW, and AW conceived of the ideas presented in this work. KK and GT provided the dataset used in pretraining \whisbert. LW implemented the model and carried out the experiments. LW, KK, GT, EGW, AW, and TIR wrote the manuscript. All authors edited the manuscript and reviewed the work. 

\bibliography{anthology,custom}

\begin{thebibliography}{21}
\expandafter\ifx\csname natexlab\endcsname\relax\def\natexlab#1{#1}\fi

\bibitem[{Baevski et~al.(2020)Baevski, Zhou, rahman Mohamed, and
  Auli}]{Baevski2020wav2vec2A}
Alexei Baevski, Henry Zhou, Abdel rahman Mohamed, and Michael Auli. 2020.
\newblock \href {https://api.semanticscholar.org/CorpusID:219966759} {wav2vec
  2.0: A framework for self-supervised learning of speech representations}.
\newblock \emph{ArXiv}, abs/2006.11477.

\bibitem[{Bain et~al.(2023)Bain, Huh, Han, and Zisserman}]{bain2023whisperx}
Max Bain, Jaesung Huh, Tengda Han, and Andrew Zisserman. 2023.
\newblock \href {http://arxiv.org/abs/2303.00747} {Whisperx: Time-accurate
  speech transcription of long-form audio}.

\bibitem[{Baltrušaitis et~al.(2017)Baltrušaitis, Ahuja, and
  Morency}]{baltrušaitis2017multimodal}
Tadas Baltrušaitis, Chaitanya Ahuja, and Louis-Philippe Morency. 2017.
\newblock \href {http://arxiv.org/abs/1705.09406} {Multimodal machine learning:
  A survey and taxonomy}.

\bibitem[{Bredin et~al.(2019)Bredin, Yin, Coria, Gelly, Korshunov, Lavechin,
  Fustes, Titeux, Bouaziz, and Gill}]{Bredin2019PyannoteAudioNB}
Herv{\'e} Bredin, Ruiqing Yin, Juan~Manuel Coria, Gregory Gelly, Pavel
  Korshunov, Marvin Lavechin, Diego Fustes, Hadrien Titeux, Wassim Bouaziz, and
  Marie-Philippe Gill. 2019.
\newblock \href {https://api.semanticscholar.org/CorpusID:207779942}
  {Pyannote.audio: Neural building blocks for speaker diarization}.
\newblock \emph{ICASSP 2020 - 2020 IEEE International Conference on Acoustics,
  Speech and Signal Processing (ICASSP)}, pages 7124--7128.

\bibitem[{Chuang et~al.(2019)Chuang, Liu, Lee, and Lee}]{chuang2019speechbert}
Yung-Sung Chuang, Chi-Liang Liu, Hung-Yi Lee, and Lin-shan Lee. 2019.
\newblock Speechbert: An audio-and-text jointly learned language model for
  end-to-end spoken question answering.
\newblock \emph{arXiv preprint arXiv:1910.11559}.

\bibitem[{Devlin et~al.(2019)Devlin, Chang, Lee, and
  Toutanova}]{devlin2019bert}
Jacob Devlin, Ming-Wei Chang, Kenton Lee, and Kristina Toutanova. 2019.
\newblock \href {http://arxiv.org/abs/1810.04805} {Bert: Pre-training of deep
  bidirectional transformers for language understanding}.

\bibitem[{Driess et~al.(2023)Driess, Xia, Sajjadi, Lynch, Chowdhery, Ichter,
  Wahid, Tompson, Vuong, Yu et~al.}]{driess2023palm}
Danny Driess, Fei Xia, Mehdi~SM Sajjadi, Corey Lynch, Aakanksha Chowdhery,
  Brian Ichter, Ayzaan Wahid, Jonathan Tompson, Quan Vuong, Tianhe Yu, et~al.
  2023.
\newblock Palm-e: An embodied multimodal language model.
\newblock \emph{arXiv preprint arXiv:2303.03378}.

\bibitem[{Frank(2023)}]{frank2023bridging}
Michael~C Frank. 2023.
\newblock \href {https://doi.org/10.31234/osf.io/qzbgx} {Bridging the data gap
  between children and large language models}.

\bibitem[{Galvez et~al.(2021)Galvez, Diamos, Ciro, Cerón, Achorn, Gopi,
  Kanter, Lam, Mazumder, and Reddi}]{galvez_peoples_2021}
Daniel Galvez, Greg Diamos, Juan Ciro, Juan~Felipe Cerón, Keith Achorn, Anjali
  Gopi, David Kanter, Maximilian Lam, Mark Mazumder, and Vijay~Janapa Reddi.
  2021.
\newblock \href {http://arxiv.org/abs/2111.09344} {The {People}'s {Speech}: {A}
  {Large}-{Scale} {Diverse} {English} {Speech} {Recognition} {Dataset} for
  {Commercial} {Usage}}.
\newblock ArXiv:2111.09344 [cs, stat].

\bibitem[{Gutmann and Hyvärinen(2010)}]{gutmann2010contrastive}
Michael Gutmann and Aapo Hyvärinen. 2010.
\newblock \href {https://proceedings.mlr.press/v9/gutmann10a.html}
  {Noise-contrastive estimation: A new estimation principle for unnormalized
  statistical models}.
\newblock In \emph{Proceedings of the Thirteenth International Conference on
  Artificial Intelligence and Statistics}, volume~9 of \emph{Proceedings of
  Machine Learning Research}, pages 297--304, Chia Laguna Resort, Sardinia,
  Italy. PMLR.

\bibitem[{Koh et~al.(2023)Koh, Fried, and Salakhutdinov}]{koh2023generating}
Jing~Yu Koh, Daniel Fried, and Ruslan Salakhutdinov. 2023.
\newblock \href {http://arxiv.org/abs/2305.17216} {Generating images with
  multimodal language models}.

\bibitem[{Lakhotia et~al.(2021)Lakhotia, Kharitonov, Hsu, Adi, Polyak, Bolte,
  Nguyen, Copet, Baevski, Mohamed et~al.}]{lakhotia2021generative}
Kushal Lakhotia, Eugene Kharitonov, Wei-Ning Hsu, Yossi Adi, Adam Polyak,
  Benjamin Bolte, Tu-Anh Nguyen, Jade Copet, Alexei Baevski, Abdelrahman
  Mohamed, et~al. 2021.
\newblock On generative spoken language modeling from raw audio.
\newblock \emph{Transactions of the Association for Computational Linguistics},
  9:1336--1354.

\bibitem[{Radford et~al.(2021)Radford, Kim, Hallacy, Ramesh, Goh, Agarwal,
  Sastry, Askell, Mishkin, Clark, Krueger, and Sutskever}]{radford2021clip}
Alec Radford, Jong~Wook Kim, Chris Hallacy, Aditya Ramesh, Gabriel Goh,
  Sandhini Agarwal, Girish Sastry, Amanda Askell, Pamela Mishkin, Jack Clark,
  Gretchen Krueger, and Ilya Sutskever. 2021.
\newblock \href {https://proceedings.mlr.press/v139/radford21a.html} {Learning
  transferable visual models from natural language supervision}.
\newblock In \emph{Proceedings of the 38th International Conference on Machine
  Learning}, volume 139 of \emph{Proceedings of Machine Learning Research},
  pages 8748--8763. PMLR.

\bibitem[{Radford et~al.(2022)Radford, Kim, Xu, Brockman, McLeavey, and
  Sutskever}]{radford2022robust}
Alec Radford, Jong~Wook Kim, Tao Xu, Greg Brockman, Christine McLeavey, and
  Ilya Sutskever. 2022.
\newblock \href {http://arxiv.org/abs/2212.04356} {Robust speech recognition
  via large-scale weak supervision}.

\bibitem[{Singh et~al.(2022)Singh, Hu, Goswami, Couairon, Galuba, Rohrbach, and
  Kiela}]{singh_flava_2022}
Amanpreet Singh, Ronghang Hu, Vedanuj Goswami, Guillaume Couairon, Wojciech
  Galuba, Marcus Rohrbach, and Douwe Kiela. 2022.
\newblock \href {http://arxiv.org/abs/2112.04482} {{FLAVA}: {A} {Foundational}
  {Language} {And} {Vision} {Alignment} {Model}}.
\newblock ArXiv:2112.04482 [cs].

\bibitem[{van~den Oord et~al.(2019)van~den Oord, Li, and
  Vinyals}]{oord2019representation}
Aaron van~den Oord, Yazhe Li, and Oriol Vinyals. 2019.
\newblock \href {http://arxiv.org/abs/1807.03748} {Representation learning with
  contrastive predictive coding}.

\bibitem[{Vaswani et~al.(2017)Vaswani, Shazeer, Parmar, Uszkoreit, Jones,
  Gomez, Kaiser, and Polosukhin}]{vaswani2017attention}
Ashish Vaswani, Noam Shazeer, Niki Parmar, Jakob Uszkoreit, Llion Jones,
  Aidan~N. Gomez, Lukasz Kaiser, and Illia Polosukhin. 2017.
\newblock \href {http://arxiv.org/abs/1706.03762} {Attention is all you need}.

\bibitem[{Warstadt and Bowman(2022)}]{warstadt2022what}
Alex Warstadt and Samuel~R Bowman. 2022.
\newblock What artificial neural networks can tell us about human language
  acquisition.
\newblock In Shalom Lappin and Jean-Philippe Bernardy, editors, \emph{Algebraic
  {Structures} in {Natural} {Language}}, pages 17--60. CRC Press.
\newblock Publisher: CRC Press.

\bibitem[{Warstadt et~al.(2023)Warstadt, Choshen, Mueller, Williams, Wilcox,
  and Zhuang}]{warstadt2023papers}
Alex Warstadt, Leshem Choshen, Aaron Mueller, Adina Williams, Ethan Wilcox, and
  Chengxu Zhuang. 2023.
\newblock \href {http://arxiv.org/abs/2301.11796} {Call for papers -- the
  babylm challenge: Sample-efficient pretraining on a developmentally plausible
  corpus}.

\bibitem[{Wu et~al.(2016)Wu, Schuster, Chen, Le, Norouzi, Macherey, Krikun,
  Cao, Gao, Macherey, Klingner, Shah, Johnson, Liu, Łukasz Kaiser, Gouws,
  Kato, Kudo, Kazawa, Stevens, Kurian, Patil, Wang, Young, Smith, Riesa,
  Rudnick, Vinyals, Corrado, Hughes, and Dean}]{wu2016googles}
Yonghui Wu, Mike Schuster, Zhifeng Chen, Quoc~V. Le, Mohammad Norouzi, Wolfgang
  Macherey, Maxim Krikun, Yuan Cao, Qin Gao, Klaus Macherey, Jeff Klingner,
  Apurva Shah, Melvin Johnson, Xiaobing Liu, Łukasz Kaiser, Stephan Gouws,
  Yoshikiyo Kato, Taku Kudo, Hideto Kazawa, Keith Stevens, George Kurian,
  Nishant Patil, Wei Wang, Cliff Young, Jason Smith, Jason Riesa, Alex Rudnick,
  Oriol Vinyals, Greg Corrado, Macduff Hughes, and Jeffrey Dean. 2016.
\newblock \href {http://arxiv.org/abs/1609.08144} {Google's neural machine
  translation system: Bridging the gap between human and machine translation}.

\bibitem[{Yasunaga et~al.(2023)Yasunaga, Aghajanyan, Shi, James, Leskovec,
  Liang, Lewis, Zettlemoyer, and Yih}]{yasunaga2023retrieval}
Michihiro Yasunaga, Armen Aghajanyan, Weijia Shi, Richard James, Jure Leskovec,
  Percy Liang, Mike Lewis, Luke Zettlemoyer, and Wen-tau Yih. 2023.
\newblock Retrieval-augmented multimodal language modeling.

\end{thebibliography}

\bibliographystyle{acl_natbib}

\newpage

\appendix

\onecolumn

\end{document}